%% file: acl_latex.tex
\documentclass[11pt]{article}

\usepackage[preprint]{acl}

\usepackage{times}
\usepackage{latexsym}

\usepackage[T1]{fontenc}

\usepackage[utf8]{inputenc}

\usepackage{microtype}

\usepackage{inconsolata}

\usepackage{graphicx}

\title{Symbolic Augmentation Closes a Canonical-Equivalence Blind Spot in Neural Fact-Checkers}

\author{Genpei Zhang\\
  University of Wisconsin-Madison \\
  }

\usepackage{amsmath,amssymb}
\usepackage{xspace}        
\usepackage{siunitx}       
\usepackage{bm}            
 
\usepackage{algorithm}
\usepackage{algpseudocode}
 
\usepackage{tikz}
\usetikzlibrary{positioning, arrows.meta, shapes.geometric, calc}
 
\usepackage{graphicx}
 
\usepackage{enumitem}
 
\usepackage{booktabs}      
\usepackage{multirow}      
 
\usepackage{seqsplit}      
 
\DeclareSIUnit{\molar}{M}
 
\newcommand{\UGV}{\textsc{UGV}\xspace}
\newcommand{\unitgraph}{\textsc{UnitGraph Verifier}\xspace}
 
\newcommand{\correctlabel}{\textsc{Correct}\xspace}
\newcommand{\uniterror}{\textsc{Unit Error}\xspace}
\newcommand{\scaleerror}{\textsc{Scale Error}\xspace}
\newcommand{\relerror}{\textsc{Relation Error}\xspace}
\newcommand{\unsupported}{\textsc{Unsupported}\xspace}
 
\newcommand{\fone}{F\textsubscript{1}\xspace}
\newcommand{\macroF}{macro-\fone\xspace}
 
\newcommand{\IC}{IC\textsubscript{50}\xspace}

\begin{document}
\maketitle

\input{sec/0_abstract}
\input{sec/1_intro}
\input{sec/2_related}
\input{sec/3_task}
\input{sec/4_method}
\input{sec/5_exp}
\input{sec/6_con}

\newpage
\input{sec/7_limi}

\bibliography{custom}

\newpage
\appendix
\input{sec/8_app}

\end{document}

%% file: sec/0_abstract.tex
%

\begin{abstract}
Large language models hallucinate numbers and units when
summarizing scientific text, a failure mode that can silently
invert a scientific claim. We recast the detection of such errors
as \emph{typed} verification: we introduce a five-class
typed-quantity error taxonomy and a 1500-item benchmark, rewritten
from PMC and arXiv sources and labeled by two independent LLM
annotators with adjudication (Krippendorff's $\alpha = 0.882$). A
ModernBERT encoder fine-tuned on this benchmark reaches
$\macroF = 0.899$, far above any off-the-shelf neural
fact-checker, yet four probes expose a sharp structural blind
spot: on canonical-equivalent rewrites of physically equivalent
quantities (e.g., \SI{95}{\degreeCelsius} $\leftrightarrow$
\SI{368.15}{\kelvin}) its accuracy collapses to $36.5\%$. We
propose \textbf{Symbolic Augmentation}, a training-time framework
that runs the modules of a symbolic verifier in reverse to
generate label-preserving augmented training data. The
augmentation lifts canonical-equivalence robustness to $98.2\%$
while slightly improving in-distribution accuracy
($\macroF$: $0.899 \to 0.902$); the augmented encoder matches a
closed-frontier LLM at no inference cost and transfers to an
external benchmark (SciFact-Open binary $\macroF$:
$0.791 \to 0.828$). Two negative results sharpen the claim:
symbolic features as auxiliary encoder inputs add nothing, and
symbolic silver labels scale negatively under teacher noise.
Together these results identify training-time augmentation as the
right integration point between symbolic and learned components.
\end{abstract}

%% file: sec/1_intro.tex
%
%

\section{Introduction}
\label{sec:intro}

LLM-based scientific writing assistants are now widely deployed:
search interfaces produce one-sentence summaries of papers
\citep{singh2024scholarqa,asai2024openscholar}, agentic pipelines
compress experimental protocols
\citep{boiko2023coscientist,bran2024chemcrow}, and authoring tools
draft method sections from raw notes \citep{liang2024monitoring}.
In all of these settings, numeric and unit hallucinations are a
uniquely dangerous failure mode: a magnitude, unit, or comparator
in the generated text disagrees with the source in a way that
changes the scientific claim, while remaining grammatically fluent.
Unlike paraphrase errors, which a careful reader can often catch
from context, a single mis-stated unit can flip the safety profile
of a drug candidate without any local linguistic cue
\citep{ji2023survey,zhang2024siren}.

Consider an evidence sentence ``Compound A inhibited EGFR with an
\IC of \SI{12}{\nano\molar}'' summarized as ``\ldots with an \IC
of \SI{12}{\milli\gram\per\milli\litre}.'' A practicing biologist
sees the problem at once: nM is a molar concentration while mg/mL
is a mass-volume concentration, and converting between them
requires the molar mass of compound A---the two summaries support
entirely different conclusions. A generic fact-check verifier,
however, will at best answer ``not entailed,'' giving no signal
about \emph{which kind} of error occurred, \emph{which span} to
repair, or whether the defect is a typo, an order-of-magnitude
slip, or a category-level confusion. We argue this task should be
modeled as \emph{typed} classification, with the verifier producing
not a binary verdict but one of five labels (\correctlabel,
\uniterror, \scaleerror, \relerror, \unsupported) that each
corresponds to a distinct symbolic axis of the quantitative claim.

To enable systematic study of this task, we construct a frozen
\num{1500}-sentence evaluation set, LLM-rewritten from real PMC
(biomedical and materials science) and arXiv (computer science)
sources, balanced across the five classes and dual-annotated by
two independent LLMs with third-party adjudication on
disagreements; Krippendorff's $\alpha = 0.882$ exceeds the standard
$0.80$ publishability threshold \citep{krippendorff1980content}.
A small probe set with a disjoint perturbation generator and a
50-item human spot-check ($94\%$ agreement) address the
self-labeling circularity concern.

On this benchmark, a fine-tuned ModernBERT encoder
\citep{warner2024modernbert,tasksource2025modernbertnli} reaches
$\macroF = 0.899$ (5-fold CV), far above any off-the-shelf neural
fact-checker we test and within $0.01$--$0.07$ of closed-frontier
LLMs at a fraction of the cost. The encoder is, by this measure,
remarkably capable---yet it exhibits a sharp \emph{structural}
blind spot. When we apply canonical-equivalent rewrites to gold
\correctlabel items---rewriting 95\,$^\circ$C as 368.15\,K, or
0.5\,mol/L as 500\,mmol/L---its accuracy on these physically
equivalent pairs drops from $96.7\%$ to $36.5\%$. Three further
probes (lexical synonyms, unit-name paraphrase, scientific
notation) localize the failure: the encoder is robust to lexical
and syntactic variation but cannot canonicalize physically
equivalent numerical surfaces.

This blind spot suggests an integration question: a symbolic
verifier canonicalizes units trivially via a registry, so can it
supplement the encoder where the encoder is blind, and at which
point in the pipeline? We answer by introducing \textbf{Symbolic
Augmentation}, a training-time framework that runs the modules of
a symbolic verifier in reverse to generate label-preserving
training data. Four augmentation families---covering unit
canonical-equivalence, comparator synonyms, unit-name paraphrase,
and scientific notation---are each derived from one verifier
module (\S\ref{sec:symaug}) and tested with a per-family probe.
Training with the canonical-equivalence family raises its probe
accuracy from $36.5\%$ to $98.2\%$ while slightly improving
in-distribution accuracy; the augmented encoder matches a
closed-frontier LLM at no inference cost and transfers to an
external benchmark. We complement these positive results with two
principled negatives---symbolic features as auxiliary inputs, and
the verifier as a silver-label teacher---that locate
training-time augmentation as the right integration point.

\paragraph{Contributions.}\mbox{}\\[-0.6em]

\textbf{(C1) Task and benchmark.} We define a five-class
typed-quantity error taxonomy with priority rules and a templated
repair specification (\S\ref{sec:task}), and release a 1500-item
benchmark rewritten from PMC and arXiv sources, dual-annotated by
two independent LLMs with third-party adjudication
($\alpha = 0.882$) and validated by a 50-item human spot-check
(\S\ref{sec:testset}).

\textbf{(C2) Symbolic Augmentation.} We localize the fine-tuned
encoder's residual failure to canonical-equivalence rewrites
through a four-family probe suite, and propose a training-time
framework that derives label-preserving augmentation rules from
the modules of a symbolic verifier, closing the blind spot and
transferring to an external benchmark (\S\ref{sec:symaug},
\S\ref{sec:ood}).

\textbf{(C3) Where symbolic information should enter.} We compare
five integration points and show that only training-time
augmentation delivers a measurable, transferable benefit:
feature-level concatenation, silver-label teaching, and
inference-time ensembling all fail (\S\ref{sec:negatives}).


%% file: sec/2_related.tex
%

\section{Related Work}
\label{sec:related}

\paragraph{Faithfulness verification.}
The dominant paradigm casts faithfulness as binary or ternary
entailment between a claim and its evidence. MiniCheck
\citep{tang2024minicheck} distills a calibrated fact-checker
matching GPT-4 at lower cost, and a line of recent systems extends
this paradigm---Bespoke-MiniCheck-7B \citep{bespoke2024minicheck},
FactCG \citep{lei2025factcg}, VeriFastScore
\citep{verifastscore2025}, and Granite Guardian
\citep{ibm2025granite}---alongside sampling- and tool-based
checkers \citep{manakul2023selfcheckgpt,chern2023factool}. A
parallel line moves beyond a single faithfulness scalar: ACUEval
\citep{wan2024acueval} types atomic units as supported, missing,
or hallucinated; FineSurE \citep{song2024finesure} predicts
sentence-level error types; and the HalluLens benchmark
\citep{bang2025hallulens} and Mu-SHROOM shared task
\citep{vazquez2025mushroom} target fine-grained hallucination.
These typed systems share our premise that a binary judgment is
information-poor, but organize errors around \emph{content-source}
relations (extraneous, missing, unsupported) rather than the
\emph{symbolic structure} of quantitative claims; the two
taxonomies are complementary. A separate line characterizes
\emph{where} fine-tuned encoders fail: probing methodology
\citep{belinkov2022probing,ribeiro2020checklist,rogers2020primer}
and input-level blind-spot prediction \citep{mi2025blindspots}
diagnose narrow failure modes that aggregate metrics hide. We
follow this tradition in \S\ref{sec:symaug} with a four-family
probe suite; concurrent work on OOD generalization in NLI
\citep{conll2025ood} similarly finds that in-distribution accuracy
can mask consequential robustness gaps.

\paragraph{Numeric and scientific claim verification.}
Verification targeted specifically at numerical and scientific
claims is more recent. Number representation in NLP has been
studied broadly \citep{thawani2021numeric,mishra2022numglue}, but
benchmarks dedicated to numerical verification have only emerged
in the past two years: QuanTemp \citep{venktesh2024quantemp} and
QuanTemp++ \citep{venktesh2025quantempplus} taxonomize claims into
statistical, temporal, comparison, and interval categories, and
the CheckThat! 2025 Task 3 \citep{venktesh2025checkthat} adopts
QuanTemp as a shared task with most submissions casting veracity
as three-class NLI \citep{heil2025dsgt,anik2025claimiq} reaching
\macroF $\leq 0.57$; VerifierFC \citep{singh2025verifierfc}
explores test-time scaling for numerical claims. SciFact
\citep{wadden2020scifact}, SciFact-Open \citep{wright2022scifact},
and SciVer \citep{wadden2021sciver} remain the canonical scientific
fact-check benchmarks, but their errors are paraphrased claims
rather than unit- or magnitude-typed perturbations. Domain-adapted
scientific LMs continue to appear---LLaMat \citep{mishra2024llamat},
HalluMat \citep{hallumat2025}, MedRAGChecker
\citep{medragchecker2026}---but as we show in
\S\ref{sec:results}, surface domain familiarity does not translate
into structural reasoning over quantities. Two distinctions
separate this entire line from our setting: existing taxonomies
partition claims by \emph{topic} (temporal vs.\ statistical),
while ours partitions them by \emph{error type} (dimension vs.\
scale vs.\ relation); and existing benchmarks do not recover
\emph{which kind} of numerical defect is present, the specific
capability our taxonomy operationalizes.

\paragraph{Symbolic verification and data augmentation.}
\unitgraph sits in the lineage of symbolic and tool-augmented
verification. Rule-based pipelines using quantulum-style extractors
\citep{quantulum3} and Pint unit registries \citep{pint} have long
been used for unit canonicalization in scientific text mining but
are seldom benchmarked as standalone verifiers. More recent work
integrates symbolic reasoning into verification: TabVer
\citep{aly2024tabver} extends natural logic with set-theoretic
arithmetic for tabular claims, and CoSineVerifier
\citep{cosineverifier2025} adds symbolic execution for
computation-heavy STEM answers. Concurrent work
\citep{seo2025verifying} similarly identifies pitfalls in fact
verifiers and explores synthetic data generation as a remedy.

A parallel line uses data augmentation to inject knowledge into
encoder training rather than at inference time. Counterfactual
data augmentation \citep{zhu2023explain,zhang2025dually} generates
label-flipping rewrites of training items, typically via learned
paraphrasers or LLM prompting
\citep{whitehouse2023llm,feng2021empirical}. Our Symbolic
Augmentation (\S\ref{sec:symaug}) is in this spirit but differs in
that its rewrite rules are derived deterministically from the
modules of a symbolic verifier---yielding stronger label
guarantees---and are \emph{label-preserving} rather than
label-flipping, the appropriate shape for closing
canonical-equivalence blind spots. The framework extends to any
symbolic verifier whose modules expose deterministic,
label-preserving rewrites.

%% file: sec/3_task.tex
%
%
%
%
%

\section{Typed-Quantity Verification}
\label{sec:task}

\begin{figure*}[t]
\centering
\includegraphics[width=\textwidth]{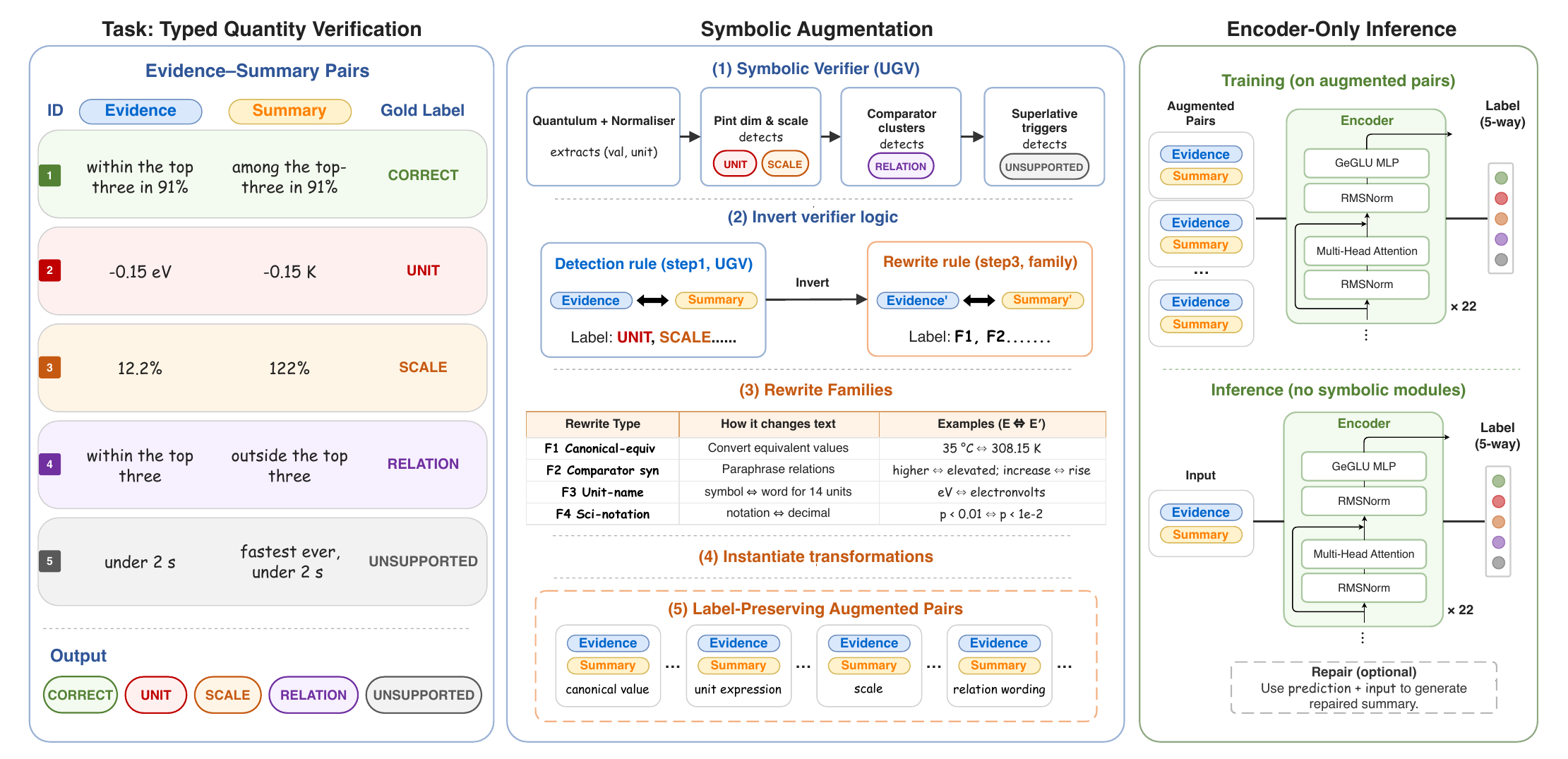}
\caption{Overview of the typed-quantity verification task and the
Symbolic Augmentation framework. \textbf{(left)} The task: each
evidence--summary pair is assigned one of five typed labels
(\correctlabel, \uniterror, \scaleerror, \relerror,
\unsupported), with one example per label. \textbf{(center)}
\textbf{Symbolic Augmentation}, our contribution: the symbolic
verifier \UGV---which extracts quantities, checks dimension and
scale, and detects comparator and superlative patterns---is run
in reverse. Each detection rule is inverted into a
label-preserving rewrite family (F1--F4), which is instantiated to
produce augmented training pairs that target the encoder's
canonical-equivalence blind spot. \textbf{(right)} At inference
time the fine-tuned encoder runs alone, with no symbolic modules;
a templated repair is optional.}
\label{fig:overview}
\end{figure*}

\subsection{Task formulation}
\label{sec:task:form}

Given an evidence sentence $e$ and a summary $s$ purporting to
restate $e$'s quantitative content, the verifier assigns a label
$y$ from one of five classes (\correctlabel, \uniterror,
\scaleerror, \relerror, \unsupported) and optionally emits a
templated repair $r$ that, applied to $s$, yields a sentence
consistent with $e$. Each non-\correctlabel class corresponds to a
distinct symbolic axis along which $s$ can deviate from $e$---unit
dimension, magnitude scale, relational structure, and
unsupported-claim addition---while \correctlabel covers everything
else, including \emph{surface variations that preserve the
underlying physical claim}, a property central to our augmentation
framework (\S\ref{sec:symaug}). The taxonomy is dataset-agnostic;
we verify transfer to an external benchmark in \S\ref{sec:ood}.

\subsection{Class definitions}
\label{sec:task:classes}

\paragraph{\correctlabel.}
$s$ is a paraphrase or faithful condensation of $e$. Crucially,
this includes \emph{canonical-equivalent rewrites}---surface forms
that canonicalize to the same physical quantity through a unit
registry or notational equivalence. For example, 95\,$^\circ$C and
368.15\,K denote the same temperature, 0.5\,mol/L and 500\,mmol/L
the same concentration, and 1.2e-9 and 0.0000000012 the same
value. Such rewrites are correct restatements, not errors, even
though their surface forms differ from the evidence.

\paragraph{\uniterror.}
$s$ contains a quantity whose unit has a \emph{different physical
dimension} from the evidence's. Example:
``\SI{12}{\nano\molar}'' $\to$ ``\SI{12}{\milli\gram\per\milli\litre}.''

\paragraph{\scaleerror.}
$s$ contains a quantity with the \emph{same physical dimension} as
the evidence's but a numerical value differing by at least one
order of magnitude ($|\log_{10}(q_s/q_e)| \geq 1$). One exception
applies: if two surface forms canonicalize to the same physical
quantity within $1\%$, the pair is a canonical-equivalent
\correctlabel restatement, not a magnitude error. Example:
``92\% conversion'' $\to$ ``9.2\% conversion.''

\paragraph{\relerror.}
$s$ contains a \emph{relational} discrepancy: (a) a comparator flip
(``higher'' $\to$ ``lower''), (b) a value bound to the wrong
entity, or (c) a within-order numerical change altering the
claim's scope (``after \SI{24}{\hour}'' $\to$
``after \SI{48}{\hour}'').

\paragraph{\unsupported.}
$s$ adds a claim or framing absent from $e$. This covers three
sub-types: superlative claims (``the highest ever reported''),
causal claims (``the dominant factor''), and significance claims
(``statistically significant''). Numeric content may still be
correct; the defect is the unsupported addition.

\subsection{Priority and evaluation}
\label{sec:task:eval}

\paragraph{Priority.}
A summary may exhibit multiple error types simultaneously. We
resolve conflicts in descending order: \unsupported, \uniterror,
\scaleerror, \relerror, \correctlabel. \unsupported dominates
typed errors because it is an epistemic shift independent of the
numeric content; the remaining ordering reflects axis precedence
(dimension before magnitude before relation). The ordering is
empirically active on $116$ of $1500$ items
($7.7\%$) and is detailed further in Appendix~\ref{app:priority}.

\paragraph{Evaluation.}
We evaluate verifiers along three axes: (i) classification
accuracy on a frozen 1500-item test set (\S\ref{sec:results});
(ii) robustness on per-family probes that apply physical-claim-%
preserving surface variations (\S\ref{sec:symaug:probes}); and
(iii) out-of-distribution transfer to an external benchmark
(\S\ref{sec:ood}). For systems that emit a templated repair, we
report repair accuracy as a secondary metric
(Appendix~\ref{app:repair}).

%% file: sec/4_method.tex
%
%

\section{Symbolic Augmentation}
\label{sec:symaug}

\subsection{The \unitgraph}
\label{sec:symaug:ugv}

\unitgraph (\UGV) is a deterministic, four-stage pipeline that
mirrors the task's symbolic structure (\S\ref{sec:task:form}):
each stage targets one error axis from \S\ref{sec:task:classes},
and a fifth implicit branch returns \correctlabel when no stage
fires. The pipeline is CPU-only with no language model at
inference time; its only non-standard dependencies are quantulum
\citep{quantulum3} for quantity extraction and Pint \citep{pint}
for the unit registry. The full pseudocode is in
Appendix~\ref{app:ugv}, and Figure~\ref{fig:overview} gives an
overview.

\paragraph{Stages 1--2: extraction and dimensional comparison.}
Quantulum maps each surface span to a tuple $(v, u, \mathit{span})$
of value, unit, and character range. A hand-curated normalizer
canonicalizes unit aliases (\texttt{M} $\to$ \texttt{mol/L};
\texttt{wt\%} $\to$ percent), scientific notation, and
context-implicit units. Aligned quantity pairs $(q_e, q_s)$ are
then checked for dimensional inequality ($\dim(q_s) \neq
\dim(q_e)$), yielding \uniterror, or for order-of-magnitude
inequality, yielding \scaleerror---subject to the $1\%$
canonical-equivalence override of \S\ref{sec:task:classes}.

\paragraph{Stages 3--4: relations and unsupported triggers.}
Stage~3 curates a small inventory of comparator clusters
(e.g., the ``high'' family: \textit{higher}, \textit{elevated},
\textit{greater}) and tags occurrences with polarity and arguments;
a matched pair is inconsistent if polarities differ (flip),
identities are swapped, or scope shifts. Stage~4 fires
\unsupported when one of a finite set of trigger phrases
(superlative, causal, significance markers; full inventory in
Appendix~\ref{app:triggers}) appears in $s$ but is absent from
$e$ and its surrounding context. \UGV reaches
$\macroF = 0.537$ on the 1500-item benchmark
(\S\ref{sec:expt:blindspot}), exceeding every off-the-shelf
neural fact-checker we test but falling well below a fine-tuned
encoder.

\subsection{The augmentation framework}
\label{sec:symaug:framework}

The four \UGV stages share a structural property we exploit:
each is a \emph{deterministic, label-preserving rewrite rule}
over surface forms. Pint's unit registry, for instance, defines
when two surface forms (95\,$^\circ$C and 368.15\,K) denote the
same physical quantity. Run in reverse, this rule generates
training data: given a gold \correctlabel item, rewriting its
quantity to a canonical-equivalent form yields a new
\correctlabel item whose label is guaranteed correct, because the
rule is label-preserving by construction.

We call the resulting framework \textbf{Symbolic Augmentation}.
It applies four augmentation families, each derived from one
\UGV module:

\begin{description}[leftmargin=*,itemsep=2pt,topsep=4pt]
\item[F1: Canonical equivalence.]
  Convert quantities to physically-equivalent units across systems
  (\SI{95}{\degreeCelsius} $\leftrightarrow$ 368.15\,K;
  mg $\leftrightarrow$ g; mM $\leftrightarrow$ M; $\%$ $\leftrightarrow$ fraction).
\item[F2: Comparator synonyms.]
  Swap comparator words on both $e$ and $s$ symmetrically
  (``higher'' $\leftrightarrow$ ``elevated'';
  ``lower'' $\leftrightarrow$ ``reduced'').
\item[F3: Unit-name paraphrase.]
  Substitute unit symbols with their word forms in both $e$ and
  $s$ (\texttt{M} $\leftrightarrow$ ``molar''; \texttt{K}
  $\leftrightarrow$ ``kelvin'').
\item[F4: Scientific notation.]
  Convert numbers between exponential and decimal forms on the
  summary side (\texttt{1.2e-9} $\leftrightarrow$
  \texttt{0.0000000012}).
\end{description}

Augmentation is applied to gold \correctlabel items only, so the
augmented training set retains the original gold-label
distribution; F1 and F4 rewrite only the summary side, while F2
and F3 rewrite both $e$ and $s$ symmetrically to preserve the
gold relation. The framework is extensible: any symbolic verifier
whose modules expose deterministic, label-preserving rewrites can
be added as a family.

\subsection{Per-family probes}
\label{sec:symaug:probes}

To diagnose where the encoder fails and to measure augmentation
effect, we construct a probe set for each family. A probe applies
the family's rewrite to the test set's \correctlabel items and
asks whether the model still labels them \correctlabel; probe
accuracy is the fraction it classifies correctly. The four probes
contain 181, 22, 30, and 25 variants for F1--F4 respectively,
derived from the 333 \correctlabel items in the test set. Rule
definitions and probe generation share code
(Appendix~\ref{app:probes}), so a probe and its corresponding
augmentation test exactly the same transformation. Probes thus
serve a dual role: a diagnostic for where fine-tuned encoders
fail (\S\ref{sec:expt:blindspot}), and a benchmark for whether
augmentation localizes its effect to the targeted failure
(\S\ref{sec:expt:augclose}).

%% file: sec/5_exp.tex
%
%
%
\begin{table*}[t]
\centering
\small
\setlength{\tabcolsep}{4.2pt}
\begin{tabular}{llcccccc}
\toprule
& \textbf{System} & \textbf{macro-$F_1$} &
\textbf{Correct} & \textbf{Unit} & \textbf{Scale} &
\textbf{Relation} & \textbf{Unsup.} \\
\midrule
\multirow{3}{*}{\textit{Trivial}}
 & predict-\correctlabel        & 0.073 & 0.363 & 0.000 & 0.000 & 0.000 & 0.000 \\
 & string-overlap               & 0.084 & 0.000 & 0.000 & 0.000 & 0.000 & 0.422 \\
 & unit-substring               & 0.225 & 0.391 & 0.437 & 0.000 & 0.000 & 0.299 \\
\midrule
\textit{Rule-based}
 & quantulum + Pint             & 0.385 & 0.373 & 0.548 & 0.708 & 0.000 & 0.299 \\
\midrule
\textit{Ours (symbolic)}
 & \UGV                         & 0.537 & 0.413 & 0.620 & 0.612 & 0.358 & 0.684 \\
\midrule
\multirow{6}{*}{\textit{Off-the-shelf}}
 & VeriFastScore                & 0.073 & 0.363 & 0.000 & 0.000 & 0.000 & 0.000 \\
 & Granite Guardian             & 0.155 & 0.000 & 0.289 & 0.000 & 0.144 & 0.343 \\
 & FactCG                       & 0.184 & 0.440 & 0.405 & 0.000 & 0.077 & 0.000 \\
 & Bespoke-MiniCheck-7B         & 0.229 & 0.353 & 0.303 & 0.000 & 0.145 & 0.343 \\
 & ModernBERT-NLI (zero-shot)   & 0.275 & 0.671 & 0.265 & 0.000 & 0.152 & 0.286 \\
 & MiniCheck                    & 0.295 & 0.714 & 0.337 & 0.000 & 0.150 & 0.275 \\
\midrule
\multirow{3}{*}{\textit{Fine-tuned}}
 & ModernBERT-FT (canonical)    & 0.899 & 0.855 & 0.923 & 0.896 & 0.892 & 0.930 \\
 & \;\;+\,F1 Symbolic Aug.      & \textbf{0.902} & 0.868 & 0.923 & 0.899 & 0.886 & 0.931 \\
 & \;\;+\,all-four Symbolic Aug.& 0.893 & 0.843 & 0.924 & 0.894 & 0.888 & 0.917 \\
\midrule
\multirow{3}{*}{\textit{Closed-frontier}}
 & DeepSeek V4-Flash            & 0.908 & 0.919 & 0.958 & 0.938 & 0.881 & 0.844 \\
 & GPT-5.5                      & 0.951 & 0.957 & 0.975 & 0.958 & 0.940 & 0.923 \\
 & Claude Opus 4.7              & 0.973 & 0.983 & 1.000 & 0.967 & 0.959 & 0.959 \\
\bottomrule
\end{tabular}
\caption{Main results on the 1500-item benchmark: macro-$F_1$ and
per-class $F_1$ over the five error types.}
\label{tab:main}
\end{table*}

\section{Experiments}
\label{sec:results}

\subsection{Setup}
\label{sec:expt:setup}
\label{sec:testset}

\paragraph{Dataset.}
We evaluate on the \num{1500}-item typed-quantity benchmark of
\S\ref{sec:task}. Items are LLM-rewritten from real PMC
(biomedical, materials science) and arXiv (computer science)
source sentences, balanced across the five classes, and
dual-annotated by two independent LLM annotators with third-party
adjudication (Krippendorff's $\alpha = 0.882$); a $50$-item human
spot-check agrees with the gold labels on $94\%$ of items.
Appendix~\ref{app:dataset} details the construction pipeline,
including source sampling, rewrite prompting, and circularity
controls.

\paragraph{Baselines.}
We compare five groups of systems. \emph{(1) Trivial baselines:}
three systems that establish the metric floor---a constant
\correctlabel predictor, a string-overlap exact-match check, and
\emph{unit-substring}, which flags a \uniterror whenever a
value-and-unit token in the summary is absent from the evidence.
\emph{(2) Rule-based:} quantulum \citep{quantulum3} paired with
the Pint unit registry \citep{pint}. \emph{(3) Our symbolic
verifier:} \UGV (\S\ref{sec:symaug:ugv}). \emph{(4) Off-the-shelf
neural fact-checkers:} MiniCheck \citep{tang2024minicheck},
Bespoke-MiniCheck-7B \citep{bespoke2024minicheck}, FactCG
\citep{lei2025factcg}, a zero-shot ModernBERT-NLI
\citep{tasksource2025modernbertnli}, Granite Guardian
\citep{ibm2025granite}, and VeriFastScore
\citep{verifastscore2025}. \emph{(5) Closed-frontier LLMs:} Claude
Opus~4.7, GPT-5.5, and DeepSeek~V4, evaluated zero-shot. Versions,
checkpoints, and invocation settings for every system are in
Appendix~\ref{app:baselines}.

\paragraph{Metrics.}
Our primary metric is the macro-averaged $F_1$ score over the
five classes. For the augmentation experiments we additionally
report \emph{probe accuracy}: the fraction of a family's probe
items (\S\ref{sec:symaug:probes}) that the model still classifies
correctly. For the out-of-distribution experiment we report
binary macro-$F_1$ on SciFact-Open. Exact metric definitions are
given in Appendix~\ref{app:metrics}.

\paragraph{Fine-tuning and augmentation.}
Our fine-tuned model is ModernBERT-NLI
\citep{warner2024modernbert,tasksource2025modernbertnli}
fine-tuned on the typed-quantity task and evaluated with 5-fold
cross-validation. Symbolic Augmentation is applied to the gold
\correctlabel training items of each fold as described in
\S\ref{sec:symaug:framework}. Training hyperparameters, hardware,
and augmentation volumes are reported in
Appendix~\ref{app:training}.

\subsection{The encoder's blind spot}
\label{sec:expt:blindspot}

Table~\ref{tab:main} reports main results, and the picture is
sharp. No off-the-shelf neural system exceeds $\macroF = 0.295$,
and the per-class columns show why: \emph{every off-the-shelf
neural system scores exactly $0.000$ on} \scaleerror. Their
\correctlabel scores vary widely ($0.00$--$0.71$)---some detect
that a quantity is being restated, but none detect that a
magnitude is wrong. The rule-based quantulum\,+\,Pint baseline
($0.385$) and \UGV ($0.537$) both outperform every neural system,
because explicit dimensional arithmetic is exactly what \scaleerror
and \uniterror require.

Fine-tuning changes the picture entirely: ModernBERT fine-tuned
on the typed-quantity task reaches $\macroF = 0.899$ (5-fold CV),
within $0.07$ of the strongest closed-frontier LLM (Claude
Opus~4.7, $0.973$) and ahead of DeepSeek~V4-Flash. A modest
encoder, given a few hundred labeled examples per class, is
competitive with frontier systems costing orders of magnitude more
per query. Yet one per-class score stands out: \correctlabel is
the encoder's \emph{weakest} class at $F_1 = 0.855$, well below
its \uniterror ($0.923$) and \unsupported ($0.930$) scores. A model
this strong should not find \correctlabel hardest---unless some
\correctlabel items share a structure it cannot handle.

The four probes of \S\ref{sec:symaug:probes} locate that
structure. The \emph{none} row of Table~\ref{tab:perfamily}
reports the canonical encoder's probe accuracies: on three of
four---comparator synonyms ($100\%$), unit-name paraphrase
($96.7\%$), scientific notation ($68\%$)---it is robust or nearly
so. On canonical-equivalence rewrites it scores only $36.5\%$:
shown a gold \correctlabel summary that states a quantity in a
physically equivalent but textually different form
(\SI{95}{\degreeCelsius} as \SI{368.15}{\kelvin}), the encoder
calls it an error nearly two-thirds of the time. This is the
structure hidden in the $0.855$ \correctlabel score:
canonical-equivalent items are a minority of \correctlabel---large
enough to depress its $F_1$, too rare for the encoder to learn
in-distribution. The failure is not noise; it concentrates on the
one transformation that requires canonicalizing a numerical
surface rather than matching lexical or syntactic patterns.

\subsection{Symbolic Augmentation closes the gap}
\label{sec:expt:augclose}

\begin{table}[t]
\centering
\small
\setlength{\tabcolsep}{6pt}
\begin{tabular}{lcc}
\toprule
\textbf{Model} & \textbf{macro-$F_1$} & \textbf{F1 probe} \\
\midrule
ModernBERT-FT (canonical)     & 0.899 & 36.5 \\
\;\;+\,F1 Symbolic Aug.       & \textbf{0.902} & \textbf{98.2} \\
\;\;+\,all-four Symbolic Aug. & 0.893 & 97.8 \\
\midrule
\multicolumn{3}{l}{\textit{Reference}} \\
\UGV                          & 0.537 & 48.1 \\
DeepSeek V4-Flash             & 0.908 & 83.4 \\
\bottomrule
\end{tabular}
\caption{Headline results, 5-fold pooled. The F1-probe column is
canonical-equivalence accuracy (\%); \UGV and a closed-frontier
LLM are shown for reference.}
\label{tab:variants}
\end{table}

\begin{table}[t]
\centering
\setlength{\tabcolsep}{6pt}
\begin{tabular}{lcccc}
\toprule
\textbf{Augmentation} & \textbf{F1} & \textbf{F2} & \textbf{F3} & \textbf{F4} \\
\midrule
none (canonical)  & 36.5 & 100 & 96.7 & 68.0 \\
\;+\,F1           & \textbf{98.9} & 100 & 96.7 & 96.0 \\
\;+\,F2           & 47.0 & 100 & 96.7 & 28.0 \\
\;+\,F3           & 33.1 & 100 & 96.7 & 32.0 \\
\;+\,F4           & 49.7 & 100 & 96.7 & \textbf{96.0} \\
\;+\,all four     & 98.9 & 100 & 100  & 96.0 \\
\bottomrule
\end{tabular}
\caption{Per-family probe accuracy (\%), single fold. Rows are
training configurations, columns the four probes; training with a
family raises its own probe (diagonal, bold).}
\label{tab:perfamily}
\end{table}

\begin{figure}[t]
\centering
\includegraphics[width=\columnwidth]{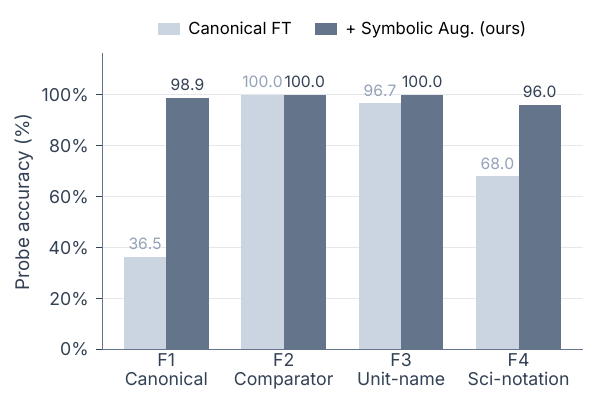}
\caption{Per-family probe accuracy before and after Symbolic
Augmentation. The canonical encoder collapses to $36.5\%$ on
canonical-equivalence rewrites (F1); augmentation closes this gap
while leaving the other probes near-saturated.}
\label{fig:blindspot}
\end{figure}

Training with the F1 (canonical-equivalence) family raises
F1-probe accuracy from $36.5\%$ to $98.2\%$ (5-fold pooled;
Table~\ref{tab:variants})---a 62-point gain that closes
essentially all of the blind spot---while macro-$F_1$ rises
slightly, $0.899 \to 0.902$. The robustness gain comes free, not
bought with in-distribution accuracy. As Table~\ref{tab:perfamily}
shows, the F4 (scientific-notation) family similarly lifts its own
probe from $68\%$ to $96\%$, while F2 and F3 yield no gain applied
alone---their probes already start near saturation---so we use
them only within the combined set. Figure~\ref{fig:blindspot}
contrasts the canonical and augmented encoders across all four
probes; training with all four families at once raises every
probe to at least $96\%$ while holding macro-$F_1$ at $0.893$,
statistically indistinguishable from baseline. We therefore carry
two variants forward: F1-augmentation as the strongest single
intervention, all-four as the broadest. We also observe
cross-family transfer---F1 alone lifts the F4 probe from $68\%$ to
$96\%$, so exposure to one numerical-surface rewrite generalizes
to another. The blind spot is not unique to small encoders:
DeepSeek V4-Flash, a closed-frontier model, scores $83.4\%$ on the
same probe---far above the canonical encoder but short of robust.
F1-augmented fine-tuning thus \emph{matches} V4-Flash on overall
macro-$F_1$ ($0.902$ vs.\ $0.908$) and \emph{exceeds} it on
canonical-equivalence robustness by nearly 15 points, while
running locally at no per-query cost.

\begin{figure}[t]
\centering
\includegraphics[width=\columnwidth]{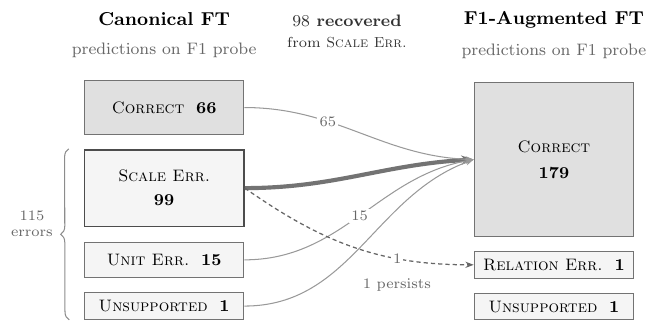}
\caption{Prediction flow on the F1 probe, all $181$ variants gold
\correctlabel. The canonical encoder (left) sends $99$ of its
$115$ errors to \scaleerror; Symbolic Augmentation (right)
recovers $114$ of the $115$.}
\label{fig:mechanism}
\end{figure}

\paragraph{Mechanism.}
Figure~\ref{fig:mechanism} traces the F1 probe at the
prediction level. All $181$ probe variants are gold \correctlabel,
yet the canonical encoder labels only $66$ of them \correctlabel.
The $115$ misclassifications are not spread evenly: $99$ of
them---$86\%$---land in \scaleerror. The encoder, shown a
canonical-equivalent rewrite such as \SI{95}{\degreeCelsius}
$\to$ \SI{368.15}{\kelvin}, most often reads it as a magnitude
that has been altered. A further $15$ land in \uniterror; almost
none reach \relerror, confirming that the failure is a
\emph{numerical-surface} confusion rather than a relational one.
Symbolic Augmentation repairs $114$ of the $115$
($99.1\%$): the rewritten \correctlabel examples teach the encoder
to read the altered surface as a restatement. A single error
persists---a milligram-to-gram rewrite the augmented model still
scores \relerror---and one previously correct item is newly
misread, for a net movement of $+113$ items into \correctlabel.

\paragraph{Out-of-distribution transfer.}
\label{sec:ood}
To test whether the effect is specific to our benchmark, we
evaluate on SciFact-Open \citep{wright2022scifact}, an external
scientific claim-verification benchmark, cast as binary
verification. Symbolic Augmentation raises binary macro-$F_1$
from $0.791$ to $0.828$ ($+0.037$) without any SciFact-Open
training data, extending a monotone ladder that runs from trivial
baselines through rule-based systems and \UGV to the fine-tuned
encoders. The improvement confirms that augmentation instills a
transferable capability---canonicalizing numerical surfaces---%
rather than a benchmark-specific pattern. The full ladder and
per-class results are in Appendix~\ref{app:ood}.

\subsection{Where symbolic information helps}
\label{sec:expt:negatives}
\label{sec:negatives}

\begin{table}[t]
\centering
\small
\setlength{\tabcolsep}{4.5pt}
\begin{tabular}{lcc}
\toprule
\textbf{Integration strategy} & \textbf{macro-$F_1$} & \textbf{$\Delta$} \\
\midrule
Canonical fine-tuning           & 0.899 & --- \\
\midrule
\multicolumn{3}{l}{\textit{Inference-time integration}} \\
\;\;Logit ensemble              & 0.895 & $-0.004$ \\
\;\;Agreement filter            & 0.888 & $-0.011$ \\
\midrule
\multicolumn{3}{l}{\textit{Feature-level integration}} \\
\;\;\UGV\ output as features     & 0.879 & $-0.020$ \\
\midrule
\multicolumn{3}{l}{\textit{Silver-label training with \UGV}} \\
\;\;$1200$ silver items          & 0.622 & $-0.277$ \\
\;\;$3298$ silver items          & 0.518 & $-0.381$ \\
\midrule
\multicolumn{3}{l}{\textit{Training-data integration (ours)}} \\
\;\;\textbf{Symbolic Augmentation} & \textbf{0.902} & $\mathbf{+0.003}$ \\
\bottomrule
\end{tabular}
\caption{Five integration strategies, grouped by where the
verifier enters the pipeline, against canonical fine-tuning. Only
training-data integration (Symbolic Augmentation) improves
macro-$F_1$; see Appendix~\ref{app:negatives}.}
\label{tab:negatives}
\end{table}

Symbolic Augmentation integrates the verifier at the
training-data layer. Four other integration points are equally
plausible a priori, yet Table~\ref{tab:negatives} shows none
works. At inference time, a logit ensemble with \UGV ($-0.004$) is
within noise, and an agreement filter---take the joint prediction
when \UGV and the encoder agree, else the encoder---reduces
mathematically to the encoder alone. At the feature level,
concatenating the verifier's output as a 12-dimensional auxiliary
vector---a one-hot \UGV prediction plus continuous module-trace
signals---costs $0.020$ macro-$F_1$: the encoder already extracts
what the features encode. Using the verifier as a silver-label
teacher is actively harmful and worsens with scale ($0.622$ at
$1200$ items, $0.518$ at $3298$), because the verifier's labels
are too noisy---it agrees with the gold label on only $53.1\%$ of
items---and inject error faster than volume compensates. Only
training-time augmentation both improves macro-$F_1$ and delivers
the 62-point probe gain, because it is the one integration point
that inherits the verifier's deterministic label guarantee without
inheriting its accuracy ceiling. A full per-method analysis is in
Appendix~\ref{app:negatives}.

%% file: sec/6_con.tex
%

\section{Conclusion}
\label{sec:conclusion}

We framed the verification of numbers and units in LLM-generated
scientific text as \emph{typed} classification and released a
five-class taxonomy and a 1500-item benchmark to study it. A
fine-tuned ModernBERT encoder reaches $\macroF = 0.899$ on this
benchmark, far beyond any off-the-shelf neural fact-checker, yet
four probes expose a sharp blind spot: it scores only $36.5\%$ on
canonical-equivalent rewrites of physically equivalent quantities.
Our main contribution, \textbf{Symbolic Augmentation}, closes this
gap by transferring the modules of a symbolic verifier into the
encoder as training-time data augmentation, raising
canonical-equivalence robustness to $98.2\%$ while slightly
improving in-distribution accuracy, matching a closed-frontier LLM
at no inference cost, and transferring to an external benchmark.
Of five ways to combine a symbolic verifier with a learned model,
only training-time augmentation delivers a transferable benefit.
We hope the benchmark and framework support further work on
bringing symbolic guarantees into learned verifiers.

%% file: sec/7_limi.tex
%

\section*{Limitations}

Three scope conditions bound our results. First, the benchmark is
built from English-language PMC and arXiv sources; the taxonomy is
defined over language-independent symbolic properties and is
deliberately dataset-agnostic, and we verify cross-dataset
transfer on SciFact-Open (\S\ref{sec:expt:augclose}), but other
languages remain future work. Second, Symbolic Augmentation
requires a symbolic verifier whose modules implement
deterministic, label-preserving rewrites; the framework is general
with respect to such verifiers, though constructing them for new
domains needs additional engineering. Third, gold labels are
LLM-produced; we mitigate the self-labeling concern with a
disjoint probe generator and a 50-item human spot-check that
agrees with the gold labels on $94\%$ of items.

%% file: sec/8_app.tex

\appendix

\section{Task Details}
\label{app:priority}

\paragraph{Priority resolution.}
A single rewritten summary can exhibit more than one error type at
once---for instance, it may both mis-state a magnitude and add an
unsupported superlative. In such cases the gold label is the
single highest-priority error present, resolved by the strict
ordering \unsupported $\succ$ \uniterror $\succ$ \scaleerror
$\succ$ \relerror $\succ$ \correctlabel. The ordering reflects two
principles. First, \unsupported ranks above every typed numeric
error because it marks an \emph{epistemic} shift independent of
the numeric content: an unsupported causal or superlative claim
misleads the reader whether or not its quantities are correct.
Second, among the three typed numeric errors the ordering follows
\emph{axis precedence}: a dimensional mismatch (\uniterror) is
more fundamental than a magnitude mismatch (\scaleerror), which is
in turn more fundamental than a relational mismatch (\relerror).
The ordering is empirically active on $116$ of the $1500$
benchmark items ($7.7\%$): in each, the rewriter was prompted for
a typed numeric error but additionally introduced a superlative,
causal, or significance phrase, and the annotators relabeled the
item \unsupported.

\paragraph{Repair.}
Beyond assigning a label, a verifier may optionally emit a
\emph{templated repair} $r$: a minimal edit that, applied to the
summary, would render it \correctlabel with respect to the
evidence. The repair template depends on the error type:

\begin{itemize}[leftmargin=1.4em,itemsep=2pt,topsep=3pt]
  \item For \uniterror and \scaleerror, the repair substitutes the
    evidence's canonical quantity for the summary's incorrect one.
  \item For \relerror, the repair restores the evidence's original
    comparator direction or entity--value binding.
  \item For \unsupported, the repair deletes the unsupported span.
\end{itemize}

\noindent
Repair is a secondary capability, not a headline metric: the
systems we focus on---the fine-tuned encoder and its augmented
variants---are classifiers and emit no repair. \UGV produces a
repair as a byproduct of its symbolic trace, since each firing
stage already localizes the offending span; we report \UGV's
repair accuracy only for completeness.
\label{app:repair}

\section{The \unitgraph Pipeline}
\label{app:ugv}

\unitgraph (\UGV) is a deterministic, four-stage verification
pipeline. Algorithm~\ref{alg:ugv} states it in full; this section
describes each stage and the symbolic checks it performs.

\begin{algorithm*}[t]
\small
\caption{\UGV verification pipeline}
\label{alg:ugv}
\begin{algorithmic}[1]
\Require evidence $e$, summary $s$
\Ensure label $y \in \{$\correctlabel, \uniterror, \scaleerror,
  \relerror, \unsupported$\}$
\State $T \gets \textsc{Triggers}(s) \setminus \textsc{Triggers}(e)$
\If{$T \neq \emptyset$} \Return \unsupported \Comment{Stage 4}
\EndIf
\State $Q_e, Q_s \gets \textsc{Extract}(e), \textsc{Extract}(s)$
\State $Q_e, Q_s \gets \textsc{Normalize}(Q_e), \textsc{Normalize}(Q_s)$
\For{$(q_e, q_s)$ in $\textsc{Align}(Q_e, Q_s)$}
  \If{$\dim(q_s) \neq \dim(q_e)$}
    \Return \uniterror \Comment{Stage 2}
  \EndIf
  \If{$|\log_{10}(q_s / q_e)| \geq 1$}
    \Return \scaleerror \Comment{Stage 2}
  \EndIf
\EndFor
\If{$\textsc{ComparatorFlip}(e, s)$ \textbf{or}
    $\textsc{EntitySwap}(e, s)$}
  \Return \relerror \Comment{Stage 3}
\EndIf
\State \Return \correctlabel
\end{algorithmic}
\end{algorithm*}

\paragraph{Stage 1: extraction and normalization.}
The pipeline first extracts every quantity mention from both the
evidence and the summary using quantulum, which returns, for each
mention, a tuple of numeric value, unit, and character span. A
hand-curated normalizer then canonicalizes these tuples: it maps
unit aliases to a single surface form (for example \texttt{M} to
\texttt{mol/L} and \texttt{wt\%} to percent), expands scientific
notation to a plain decimal, and resolves context-implicit units.
Normalization ensures that two physically equivalent quantities
written differently are compared on equal footing in the later
stages.

\paragraph{Stage 2: dimensional and magnitude checks.}
After normalization, each summary quantity $q_s$ is aligned with
its corresponding evidence quantity $q_e$ by matching entity head
and modified property. For each aligned pair, \UGV applies two
checks. The \emph{dimensional check} compares physical dimensions:
if
\[
  \dim(q_s) \neq \dim(q_e),
\]
the pair is a dimensional mismatch and the pipeline returns
\uniterror. Otherwise the \emph{magnitude check} compares
canonicalized values on a logarithmic scale: if
\[
  \bigl|\log_{10}(q_s / q_e)\bigr| \geq 1,
\]
the two values differ by at least one order of magnitude and the
pipeline returns \scaleerror. The magnitude check is suppressed
when the two surface forms canonicalize to the same physical
quantity within $1\%$, which prevents a canonical-equivalent
restatement from being misreported as a magnitude error.

\paragraph{Stage 3: relational checks.}
If no quantity pair triggers a dimensional or magnitude error,
\UGV checks relational structure. The \textsc{ComparatorFlip}
test detects a reversed comparison: it tags comparator words in
both sentences with a polarity drawn from curated clusters (the
``high'' cluster contains \emph{higher}, \emph{elevated},
\emph{greater}, and so on; the ``low'' cluster its antonyms) and
fires when matched comparators carry opposite polarity. The
\textsc{EntitySwap} test detects a quantity bound to the wrong
entity. If either test fires, the pipeline returns \relerror.

\paragraph{Stage 4: unsupported-claim triggers.}
Although highest in priority and therefore executed first, this
stage checks whether the summary introduces a claim absent from
the evidence. \UGV maintains an inventory of trigger phrases in
three groups:

\begin{description}[leftmargin=1.4em,itemsep=2pt,topsep=3pt]
  \item[Superlative] highest, lowest, largest, smallest, best,
    worst, unprecedented, record.
  \item[Causal] causes, leads to, results in, due to, responsible
    for, drives, dominant factor.
  \item[Significance] statistically significant, significant
    improvement, significantly, $p < 0.05$, demonstrates.
\end{description}

\noindent
The pipeline computes the set of trigger phrases occurring in the
summary but not in the evidence; if this set is non-empty, it
returns \unsupported. Matching is case-insensitive and operates
over lemmatized tokens, so that inflectional variants
(\emph{cause}, \emph{causes}, \emph{caused}) are treated alike.
The five significance-group phrases are reused, as the only
unsupported triggers, by the unit-substring and quantulum
baselines described in Appendix~\ref{app:baselines}.
\label{app:triggers}

\section{Probe Construction}
\label{app:probes}

Each augmentation family $F_i$ is defined by a set of
deterministic, label-preserving rewrite rules over quantity
surfaces. The per-family \emph{probe} is built directly from these
rules, so that a probe and its corresponding training augmentation
exercise exactly the same transformation---they call the same
rewrite code.

Probe construction proceeds as follows. We take the $333$ test-set
items whose gold label is \correctlabel. For each item, we apply
the family's rewrite rules to the summary, producing one or more
\emph{variants} in which a quantity has been re-expressed in a
physically equivalent but textually different form. Because every
rewrite rule is label-preserving by construction, each variant
still has gold label \correctlabel. The probe for family $F_i$ is
the resulting set of variants $V_i$, and the model is scored on
whether it continues to assign them \correctlabel.

The four probes contain
\[
  |V_1| = 181, \quad |V_2| = 22, \quad
  |V_3| = 30, \quad |V_4| = 25
\]
variants respectively. The counts differ because a family can
only rewrite an item that contains an applicable quantity: F4
(scientific-notation rewriting), for example, applies only to
items whose summary contains a number that can be re-expressed in
scientific notation, and comparatively few \correctlabel items
qualify.

\section{Dataset Construction}
\label{app:dataset}

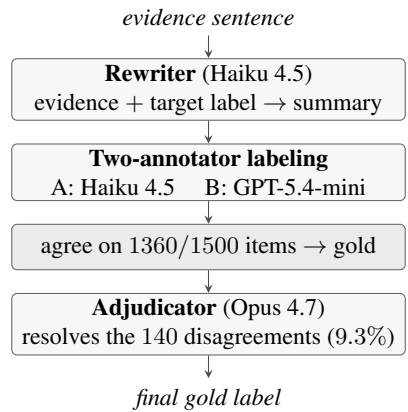
\begin{figure}[t]
\centering
\begin{tikzpicture}[
  node distance=3mm,
  every node/.style={font=\footnotesize},
  box/.style={
    rectangle, rounded corners=2pt, draw=black!60,
    line width=0.4pt, minimum width=52mm, minimum height=6.5mm,
    align=center, inner sep=2pt, fill=black!3},
  io/.style={font=\footnotesize\itshape, align=center},
  dec/.style={
    rectangle, rounded corners=2pt, draw=black!60,
    line width=0.4pt, minimum width=52mm, minimum height=6mm,
    align=center, inner sep=2pt, fill=black!8},
  arr/.style={->, >=stealth, line width=0.4pt, draw=black!70}
]
\node[io] (src) {evidence sentence};
\node[box, below=of src] (rw) {%
  \textbf{Rewriter} (Haiku 4.5)\\[1pt]
  evidence $+$ target label $\to$ summary};
\node[box, below=of rw] (ann) {%
  \textbf{Two-annotator labeling}\\[1pt]
  A: Haiku 4.5 \quad B: GPT-5.4-mini};
\node[dec, below=of ann] (dec) {%
  agree on $1360/1500$ items $\to$ gold};
\node[box, below=of dec] (adj) {%
  \textbf{Adjudicator} (Opus 4.7)\\[1pt]
  resolves the $140$ disagreements ($9.3\%$)};
\node[io, below=of adj] (gold) {final gold label};
\draw[arr] (src) -- (rw);
\draw[arr] (rw) -- (ann);
\draw[arr] (ann) -- (dec);
\draw[arr] (dec) -- (adj);
\draw[arr] (adj) -- (gold);
\end{tikzpicture}
\caption{The annotation pipeline. Each rewritten summary is
labeled independently by two annotators from different model
families; the $9.3\%$ of items on which they disagree are
resolved by an adjudicator. Krippendorff's $\alpha$ is computed
over the two annotators only.}
\label{fig:annotation}
\end{figure}

This section details the four-step construction pipeline:
source-sentence sampling, LLM rewriting, two-annotator labeling
with adjudication, and a human spot-check.
Figure~\ref{fig:annotation} summarizes the labeling flow.

\paragraph{Source sampling.}
Evidence sentences are drawn from three scientific domains, each
contributing roughly one third of the benchmark:

\begin{itemize}[leftmargin=1.4em,itemsep=3pt,topsep=3pt]
  \item \textbf{PMC biomedical} ($484$ sentences). We retrieve
    approximately $120$ open-access articles through the NCBI
    E-utilities API, using a query targeting concentration and
    pharmacokinetic terminology (for example \texttt{mg/mL},
    \texttt{\textmu M}, \texttt{IC50}, \emph{pharmacokinetic}),
    then extract sentences at the sentence level.
  \item \textbf{PMC materials science} ($507$ sentences). A
    parallel retrieval of roughly $120$ open-access articles uses
    a query targeting battery and photovoltaic terminology (for
    example \emph{perovskite}, \emph{cathode}, \texttt{mAh/g},
    \emph{photocurrent}).
  \item \textbf{arXiv computer science} ($509$ sentences).
    Sentences are retrieved through the Semantic Scholar and
    arXiv APIs with queries targeting evaluation and benchmarking
    terminology.
\end{itemize}

\noindent
A candidate sentence is retained only if quantulum can extract at
least one quantity from it, ensuring every evidence sentence has
quantitative content. Evidence sentences are further restricted to
at most three quantity tuples, which keeps quantity alignment
unambiguous in later processing. The $1500$ benchmark items derive
from $741$ unique evidence sentences, so each evidence yields
roughly two rewritten summaries on average.

\paragraph{Rewriting.}
Each evidence sentence is rewritten into candidate summaries by
\seqsplit{claude-haiku-4-5} (maximum $500$ output tokens). The
system prompt supplies the complete five-class definitions
together with per-class strategy guidelines describing how to
construct each error type; the user prompt supplies one evidence
sentence and one target label, and requests a single fluent
summary that, compared against the evidence, exhibits the targeted
class. The model returns the summary wrapped in explicit
delimiters. We rewrite each evidence against multiple target
labels, cycling through classes until each class has accumulated
$300$ candidates, for $1500$ candidates in total.

\paragraph{Annotation and adjudication.}
Every candidate summary is then independently labeled by two
annotators drawn from \emph{different model families}, which is
central to avoiding a self-labeling artifact:

\begin{itemize}[leftmargin=1.4em,itemsep=3pt,topsep=3pt]
  \item \textbf{Annotator A}: \seqsplit{claude-haiku-4-5}
    (Anthropic).
  \item \textbf{Annotator B}: \seqsplit{gpt-5.4-mini} (OpenAI).
  \item \textbf{Adjudicator}: \seqsplit{claude-opus-4-7}
    (Anthropic), invoked only on items where A and B disagree.
\end{itemize}

\noindent
Each annotator receives the evidence, the candidate summary, and
the five-class definitions, and returns a single label. The two
annotators agree on $1360$ of $1500$ items; on the remaining
$140$ ($9.3\%$) the adjudicator assigns the final label.
Inter-annotator agreement is measured by Krippendorff's $\alpha$
computed over annotators A and B only:
\[
  \alpha = 1 - \frac{D_o}{D_e} = 0.882,
\]
where $D_o$ and $D_e$ are the observed and expected disagreement.
The adjudicator does not enter this computation. Because annotator
B belongs to a different model family than the rewriter
(\seqsplit{claude-haiku-4-5}), this $\alpha$ does not reflect a
model scoring its own output, and the value comfortably exceeds
the conventional $0.80$ threshold for reliable annotation.

\paragraph{Gold label distribution.}
The rewriter targets a balanced $300$ items per class, but
adjudication shifts the final distribution, because the priority
rule of Appendix~\ref{app:priority} relabels some items. The gold
distribution is:

\begin{center}
\small
\begin{tabular}{lc}
\toprule
\textbf{Class} & \textbf{Gold count} \\
\midrule
\correctlabel  & 333 \\
\uniterror     & 236 \\
\scaleerror    & 240 \\
\relerror      & 290 \\
\unsupported   & 401 \\
\midrule
\textbf{Total} & \textbf{1500} \\
\bottomrule
\end{tabular}
\end{center}

\noindent
The final distribution departs from the balanced $300$-per-class
target for two reasons: the priority rule of
Appendix~\ref{app:priority} relabels $116$ items to \unsupported,
and annotators sometimes judged a rewrite to realize a different
class than the one targeted. Both effects move mass toward
\unsupported and \correctlabel and away from \uniterror and
\scaleerror. The three source domains remain near-balanced,
contributing $484$, $507$, and $509$ items.

\paragraph{Human spot-check.}
To validate the LLM gold labels against a human standard, one of
the authors---blind to all model predictions and to the paper's
analysis---independently labeled a random $50$-item sample. The
human label matches the adjudicated gold label on $47$ of $50$
items ($94\%$). All three disagreements are
\scaleerror-versus-\relerror boundary cases involving
sub-ten-fold numerical differences, a region in which the LLM
annotators also reported low confidence; they are genuine
borderline cases rather than annotation errors.

\section{Baseline Configurations}
\label{app:baselines}

We evaluate five groups of baseline systems. This section
specifies each one: its checkpoint or definition, its native
output, and---for systems whose native output is not already in
our label space---how that output is routed to a five-class label.

\paragraph{Trivial baselines.}
These establish the floor of the metric:

\begin{description}[leftmargin=1.4em,itemsep=3pt,topsep=3pt]
  \item[predict-\correctlabel] returns \correctlabel for every
    item, unconditionally. It measures the score obtainable with
    no modeling at all, and coincides numerically with a
    majority-class predictor, since \correctlabel is the majority
    class of the development split.
  \item[string-overlap] returns \correctlabel if the evidence and
    summary are identical after whitespace trimming, and
    \unsupported otherwise. It uses exact matching, with no
    similarity threshold.
  \item[unit-substring] extracts each value-and-unit token from
    the summary with a regular expression. It first checks for the
    five significance-group trigger phrases
    (Appendix~\ref{app:triggers}) and returns \unsupported if any
    is present; otherwise it returns \uniterror if any extracted
    token is absent from the evidence, and \correctlabel if all
    tokens are present.
\end{description}

\paragraph{Rule-based baseline.}
\begin{description}[leftmargin=1.4em,itemsep=3pt,topsep=3pt]
  \item[quantulum~$+$~Pint] extracts quantity tuples from both
    sentences with quantulum, position-aligns them, and applies
    the Stage~2 checks of Appendix~\ref{app:ugv}: a dimensional
    mismatch yields \uniterror and an order-of-magnitude
    difference yields \scaleerror. It performs no comparator or
    entity-binding analysis, which is the capability \UGV adds in
    Stage~3.
\end{description}

\paragraph{Our symbolic verifier.}
\begin{description}[leftmargin=1.4em,itemsep=3pt,topsep=3pt]
  \item[\UGV] is the full four-stage pipeline of
    Appendix~\ref{app:ugv}. It uses neither task supervision nor a
    language model at inference time.
\end{description}

\paragraph{Off-the-shelf neural fact-checkers.}
Each of the following emits a binary or ternary entailment
decision natively. We list each system with its checkpoint and
native output:

\begin{description}[leftmargin=1.4em,itemsep=3pt,topsep=3pt]
  \item[MiniCheck] (\seqsplit{lytang/MiniCheck-Flan-T5-Large}): a
    Flan-T5 model emitting a ``Yes''/``No'' supported decision.
  \item[FactCG] (\seqsplit{yaxili96/FactCG-DeBERTa-v3-Large}): a
    DeBERTa-v3 classifier whose positive label denotes
    ``supported.''
  \item[ModernBERT-NLI, zero-shot]
    (\seqsplit{tasksource/ModernBERT-large-nli}): a three-way NLI
    model; we read the ``entailment'' class as supported.
  \item[Bespoke-MiniCheck-7B]
    (\seqsplit{bespokelabs/Bespoke-MiniCheck-7B}): a 7B
    instruction-tuned model emitting a ``Yes''/``No'' decision.
  \item[Granite Guardian]
    (\seqsplit{ibm-granite/granite-guardian-3.3-8b}): an 8B
    groundedness judge emitting a ``Yes''/``No'' decision via its
    chat template.
  \item[VeriFastScore] (\seqsplit{lytang/VeriFastScore-8B}): an
    8B high-throughput verifier that emits a binary supported
    decision.
\end{description}

\noindent
\textbf{Label routing.} Because these systems emit only a support
decision, we route their output into the five-class space with a
fixed rule. A ``supported'' or ``entailment'' output maps directly
to \correctlabel. For any non-supported output, we apply the
following cascade:

\begin{enumerate}[leftmargin=1.6em,itemsep=2pt,topsep=3pt]
  \item if the summary carries an unsupported trigger phrase
    absent from the evidence, return \unsupported;
  \item else if a comparator flip is detected between summary and
    evidence, return \relerror;
  \item else return \uniterror as the default.
\end{enumerate}

\noindent
This cascade has a structural consequence: it can never output
\scaleerror. This is why every off-the-shelf neural baseline
scores exactly $0.000$ on \scaleerror in Table~\ref{tab:main}---a
limitation of the binary-entailment paradigm, not a tuning
artifact.

\paragraph{Closed-frontier LLMs.}
We evaluate three closed-weight models---Claude Opus~4.7, GPT-5.5,
and DeepSeek V4-Flash---accessed through their respective APIs.
All three are run zero-shot with chain-of-thought prompting and
greedy decoding (temperature $0$). The system prompt contains the
full five-class definitions and four in-prompt example triples,
each a (evidence, summary, label) example for a distinct class;
the user prompt supplies the evidence and summary and requests a
label. Output token budgets are $512$ for Claude and GPT and
$1024$ for DeepSeek V4-Flash, which requires extra space for
reasoning tokens. All three are evaluated on the full $1500$-item
benchmark.

\section{Metrics}
\label{app:metrics}

This section gives the precise definition of every metric used in
the paper.

\paragraph{Macro-$F_1$.}
For each class $c$, let $\mathrm{TP}_c$, $\mathrm{FP}_c$, and
$\mathrm{FN}_c$ denote true positives, false positives, and false
negatives. The per-class precision and recall are
\[
  P_c = \frac{\mathrm{TP}_c}{\mathrm{TP}_c + \mathrm{FP}_c},
  \qquad
  R_c = \frac{\mathrm{TP}_c}{\mathrm{TP}_c + \mathrm{FN}_c},
\]
and the per-class $F_1$ is their harmonic mean,
\[
  F_{1,c} = \frac{2\,P_c\,R_c}{P_c + R_c}.
\]
The primary metric, macro-$F_1$, is the unweighted mean over the
five classes,
\[
  F_1^{\mathrm{macro}}
  = \frac{1}{5} \sum_{c \in \mathcal{C}} F_{1,c},
\]
where $\mathcal{C}$ ranges over the five classes \correctlabel,
\uniterror, \scaleerror, \relerror, and \unsupported. The
unweighted mean gives the rare and common classes equal weight, so
a system cannot score well merely by serving the majority class.

\paragraph{Probe accuracy.}
For augmentation family $F_i$ with probe variant set $V_i$
(Appendix~\ref{app:probes}), every variant $v \in V_i$ has gold
label \correctlabel. Probe accuracy is the fraction the model
$\hat{y}$ still classifies \correctlabel,
\[
  \mathrm{acc}_{F_i}
  = \frac{1}{|V_i|} \sum_{v \in V_i}
    \mathbb{1}\!\left[\, \hat{y}(v) = \correctlabel \,\right],
\]
where $\mathbb{1}[\cdot]$ is the indicator function. Since every
probe variant is a label-preserving rewrite of a genuinely
\correctlabel item, probe accuracy is exactly the model's recall
on that family's surface transformation.

\paragraph{Out-of-distribution metric.}
SciFact-Open provides binary SUPPORT/CONTRADICT labels. To apply
our five-class models to it, we fold each prediction $y$ with the
mapping
\[
  \phi(y) =
  \begin{cases}
    \text{SUPPORT}    & \text{if } y = \correctlabel,\\[3pt]
    \text{CONTRADICT} & \text{otherwise},
  \end{cases}
\]
that is, the four error classes all map to CONTRADICT. We then
report binary macro-$F_1$, the unweighted mean of the SUPPORT and
CONTRADICT $F_1$ scores. The out-of-distribution evaluation is
detailed in Appendix~\ref{app:ood}.

\section{Training Details}
\label{app:training}

\paragraph{Model and optimization.}
The fine-tuned model uses
\seqsplit{tasksource/ModernBERT-large-nli} ($400$M parameters) as
its backbone. The pretrained three-class NLI head is discarded and
replaced with a randomly initialized five-class linear classifier;
the size mismatch between the old and new heads is handled by
allowing mismatched checkpoint sizes at load time. The full
optimization configuration is:

\begin{itemize}[leftmargin=1.4em,itemsep=2pt,topsep=3pt]
  \item \textbf{Optimizer}: AdamW, learning rate
    $2 \times 10^{-5}$, weight decay $0.01$.
  \item \textbf{Schedule}: linear warmup over the first $10\%$ of
    steps, followed by linear decay.
  \item \textbf{Batch size}: $8$, limited by GPU memory at
    sequence length $512$.
  \item \textbf{Epochs}: $3$, with evaluation at the end of
    training and no early stopping.
  \item \textbf{Sequence length}: $512$ tokens, with dynamic
    padding within each batch.
  \item \textbf{Loss}: cross-entropy.
  \item \textbf{Gradient clipping}: maximum norm $1.0$.
  \item \textbf{Precision}: float32, with no mixed precision.
\end{itemize}

\paragraph{Hardware.}
Each fold is trained on a single NVIDIA RTX~3090 GPU ($24$GB). A
single fold completes in two to three minutes, and the full
five-fold run completes in twelve to fifteen minutes of
wall-clock time.

\paragraph{Cross-validation.}
We use stratified five-fold cross-validation, stratified by gold
label, with $1200$ training and $300$ test items per fold. We
report two evaluation modes. The \emph{5-fold pooled} mode---used
for all headline numbers---pools the out-of-fold predictions
across all five folds and computes macro-$F_1$ on the full $1500$
items. The \emph{single-fold} mode uses one fixed $80/20$
stratified split. Pooled numbers are more stable, since per-fold
single-split numbers fluctuate with the particular split.

\paragraph{Augmentation volume.}
Symbolic Augmentation is applied only to the gold \correctlabel
items of each training fold, so the label distribution of the
augmented training set matches the original. F1 augmentation adds
approximately $155$ canonical-equivalent items per fold (its
rewrite rules applied to the fold's \correctlabel items, capped at
three new items per source item), giving a combined training set
of roughly $1350$ items. All-four augmentation adds approximately
$206$ items per fold (capped at four per source item across the
four families), giving roughly $1406$ items.

\section{Negative-Result Setups}
\label{app:negatives}

This section specifies the configuration of each integration point
compared in Table~\ref{tab:negatives}.

\paragraph{Auxiliary symbolic features.}
This setup concatenates the symbolic verifier's output to the
encoder as an auxiliary input. The \UGV output is encoded as a
$12$-dimensional vector: a $5$-dimensional one-hot of the \UGV
predicted label, plus a $7$-dimensional continuous trace of
intermediate signals---the log-scaled evidence and summary
quantity counts, a dimensional-mismatch flag, a comparator-flip
flag, a superlative-trigger flag, a scale-gap log-ratio clamped to
$[-3, 3]$, and an entity-swap flag. Architecturally, the
ModernBERT encoder output is mean-pooled over non-padding tokens,
concatenated with a $64$-dimensional learned projection of the
auxiliary vector, and passed through a two-layer classification
head. Training otherwise matches the canonical fine-tuning
configuration of Appendix~\ref{app:training}. To isolate the
contribution of the auxiliary features, we train two variants that
are identical except that one receives the true feature vector and
the other receives a zeroed vector.

\paragraph{Silver-label teachers.}
This setup uses \UGV as a teacher that labels training data, which
is then used to train the encoder. We evaluate it at two scales:

\begin{itemize}[leftmargin=1.4em,itemsep=3pt,topsep=3pt]
  \item \textbf{$n = 1200$}: the gold labels of the canonical
    training split are replaced with \UGV's predicted labels.
    \UGV agrees with the gold label on $53.1\%$ of these items, so
    the silver set is substantially noisier than gold.
  \item \textbf{$n = 3298$}: in addition, five new perturbations
    per evidence are generated by \seqsplit{gpt-5.4-mini} with the
    same five-class taxonomy prompt as the main rewriter, and each
    is labeled by \UGV. After deduplication against existing test
    summaries, $3298$ silver items remain.
\end{itemize}

\noindent
In both cases the silver set replaces the gold training set
entirely; the encoder is otherwise trained as in
Appendix~\ref{app:training}.

\paragraph{Inference-time logit ensemble.}
This setup keeps the canonical encoder unchanged and combines its
output with \UGV's at inference time. Since \UGV is rule-based and
emits a hard label rather than a distribution, its prediction is
represented as a one-hot vector. The ensemble distribution is the
convex combination
\[
  p = (1 - \alpha)\, p_{\mathrm{FT}}
      + \alpha \cdot \mathrm{onehot}(\hat{y}_{\UGV}),
\]
where $p_{\mathrm{FT}}$ is the encoder's softmax distribution and
$\alpha = 0.25$. The final prediction is $\arg\max_c p_c$.

\paragraph{Agreement filter.}
This setup uses \UGV's prediction only as a gate. For each item,
if \UGV and the encoder predict the same label, that label is
returned; if they disagree, the encoder's prediction is returned.
Formally the output is
\[
  \hat{y} =
  \begin{cases}
    \hat{y}_{\UGV}
      & \text{if } \hat{y}_{\UGV} = \hat{y}_{\mathrm{FT}},\\[3pt]
    \hat{y}_{\mathrm{FT}}
      & \text{otherwise.}
  \end{cases}
\]
In the first branch $\hat{y}_{\UGV} = \hat{y}_{\mathrm{FT}}$, so
the output equals $\hat{y}_{\mathrm{FT}}$; in the second branch it
equals $\hat{y}_{\mathrm{FT}}$ by definition. The filter therefore
returns $\hat{y}_{\mathrm{FT}}$ on every item and is mathematically
identical to the encoder alone, which is why its macro-$F_1$
exactly matches the canonical baseline.

\section{Out-of-Distribution Evaluation Details}
\label{app:ood}

\paragraph{Evaluation set.}
We evaluate out-of-distribution transfer on SciFact-Open
\citep{wright2022scifact}. We load the test split and keep the
$206$ items carrying a SUPPORT or CONTRADICT label---$105$ and
$101$ respectively---discarding items labeled
not-enough-information. No further filtering by domain, quantity
content, or sentence length is applied. Both fine-tuned models are
evaluated zero-shot: they are trained only on the typed-quantity
benchmark and never observe SciFact-Open data or its label space,
so their five-class predictions are folded to binary using the
mapping $\phi$ of Appendix~\ref{app:metrics}.

\paragraph{Ladder of systems.}
Table~\ref{tab:ood-ladder} places the two fine-tuned models on a
ladder with the same rule-based and symbolic systems used in the
main experiments. Performance increases monotonically along the
ladder: from the trivial baselines, through the rule-based systems
and \UGV, to the canonical fine-tuned encoder, and finally to the
F1-augmented encoder. Symbolic Augmentation adds $+0.037$ binary
macro-$F_1$ on top of canonical fine-tuning, entirely through
zero-shot transfer.

\begin{table}[t]
\centering
\small
\begin{tabular}{lcc}
\toprule
\textbf{System} & \textbf{macro-$F_1$} & \textbf{Acc.} \\
\midrule
predict-\correctlabel        & 0.338 & 0.510 \\
string-overlap               & 0.329 & 0.490 \\
unit-substring               & 0.390 & 0.534 \\
quantulum $+$ Pint           & 0.427 & 0.549 \\
\UGV                         & 0.455 & 0.539 \\
\midrule
ModernBERT-FT (canonical)    & 0.791 & 0.796 \\
\;\;$+$\,F1 Symbolic Aug.    & \textbf{0.828} & \textbf{0.830} \\
\bottomrule
\end{tabular}
\caption{Out-of-distribution results on SciFact-Open
($n = 206$), for the same rule-based and fine-tuned systems as the
main experiments.}
\label{tab:ood-ladder}
\end{table}

\paragraph{Per-class breakdown.}
Table~\ref{tab:ood-perclass} breaks the two fine-tuned models down
by class. The improvement is concentrated in SUPPORT recall, which
rises from $0.629$ to $0.714$, a gain of $8.5$ points. The
augmented encoder recognizes more SciFact-Open claims that
paraphrase the evidence with canonical-equivalent units as
genuinely supporting the evidence, rather than mistaking them for
relational errors or unsupported assertions. Concretely, of the
$11$ items the augmented model newly labels \correctlabel---hence
SUPPORT---seven had been assigned \relerror and four \unsupported
by the canonical model. This is the same capability the F1 probe
measures in distribution, now visible on an external benchmark.

\begin{table}[t]
\centering
\small
\setlength{\tabcolsep}{4pt}
\begin{tabular}{llccc}
\toprule
\textbf{Model} & \textbf{Class} & \textbf{P} & \textbf{R} & \textbf{$F_1$} \\
\midrule
\multirow{2}{*}{canonical FT}
 & SUPPORT     & 0.957 & 0.629 & 0.759 \\
 & CONTRADICT  & 0.715 & 0.970 & 0.824 \\
\midrule
\multirow{2}{*}{$+$\,F1 Aug.}
 & SUPPORT     & 0.938 & 0.714 & 0.811 \\
 & CONTRADICT  & 0.762 & 0.951 & 0.846 \\
\bottomrule
\end{tabular}
\caption{Per-class out-of-distribution results on SciFact-Open.
Symbolic Augmentation lifts SUPPORT recall by $8.5$ points, the
main source of the macro-$F_1$ gain.}
\label{tab:ood-perclass}
\end{table}

%% file: custom.bib
@inproceedings{singh2024scholarqa,
  title={Ai2 scholar qa: Organized literature synthesis with attribution},
  author={Singh, Amanpreet and Chang, Joseph Chee and Haddad, Dany and Naik, Aakanksha and Hwang, Jena D and Kinney, Rodney and Weld, Daniel S and Downey, Doug and Feldman, Sergey},
  booktitle={Proceedings of the 63rd Annual Meeting of the Association for Computational Linguistics (Volume 3: System Demonstrations)},
  pages={513--523},
  year={2025}
}

@article{asai2024openscholar,
  title={Openscholar: Synthesizing scientific literature with retrieval-augmented lms},
  author={Asai, Akari and He, Jacqueline and Shao, Rulin and Shi, Weijia and Singh, Amanpreet and Chang, Joseph Chee and Lo, Kyle and Soldaini, Luca and Feldman, Sergey and D'arcy, Mike and others},
  journal={arXiv preprint arXiv:2411.14199},
  year={2024}
}

@article{boiko2023coscientist,
  title={Autonomous chemical research with large language models},
  author={Boiko, Daniil A and MacKnight, Robert and Kline, Ben and Gomes, Gabe},
  journal={Nature},
  volume={624},
  number={7992},
  pages={570--578},
  year={2023},
  publisher={Nature Publishing Group UK London}
}

@article{bran2024chemcrow,
  title={Augmenting large language models with chemistry tools},
  author={M. Bran, Andres and Cox, Sam and Schilter, Oliver and Baldassari, Carlo and White, Andrew D and Schwaller, Philippe},
  journal={Nature machine intelligence},
  volume={6},
  number={5},
  pages={525--535},
  year={2024},
  publisher={Nature Publishing Group UK London}
}

@article{liang2024monitoring,
  title={Monitoring ai-modified content at scale: A case study on the impact of chatgpt on ai conference peer reviews},
  author={Liang, Weixin and Izzo, Zachary and Zhang, Yaohui and Lepp, Haley and Cao, Hancheng and Zhao, Xuandong and Chen, Lingjiao and Ye, Haotian and Liu, Sheng and Huang, Zhi and others},
  journal={arXiv preprint arXiv:2403.07183},
  year={2024}
}

@article{ji2023survey,
  title={Survey of hallucination in natural language generation},
  author={Ji, Ziwei and Lee, Nayeon and Frieske, Rita and Yu, Tiezheng and Su, Dan and Xu, Yan and Ishii, Etsuko and Bang, Ye Jin and Madotto, Andrea and Fung, Pascale},
  journal={ACM computing surveys},
  volume={55},
  number={12},
  pages={1--38},
  year={2023},
  publisher={ACM New York, NY}
}

@article{zhang2024siren,
  title={Siren’s song in the AI ocean: A survey on hallucination in large language models. arXiv 2023},
  author={Zhang, Yue and Li, Yafu and Cui, Leyang and Cai, Deng and Liu, Lemao and Fu, Tingchen and Huang, Xinting and Zhao, Enbo and Zhang, Yu and Chen, Yulong and others},
  journal={arXiv preprint arXiv:2309.01219},
  year={2023}
}

@inproceedings{tang2024minicheck,
  title={Minicheck: Efficient fact-checking of llms on grounding documents},
  author={Tang, Liyan and Laban, Philippe and Durrett, Greg},
  booktitle={Proceedings of the 2024 Conference on Empirical Methods in Natural Language Processing},
  pages={8818--8847},
  year={2024}
}

@misc{bespoke2024minicheck,
  title        = {Bespoke-{M}ini{C}heck-7{B}},
  author       = {{Bespoke Labs}},
  year         = {2024},
  howpublished = {\url{https://huggingface.co/bespokelabs/Bespoke-MiniCheck-7B}}
}

@inproceedings{manakul2023selfcheckgpt,
  title={Selfcheckgpt: Zero-resource black-box hallucination detection for generative large language models},
  author={Manakul, Potsawee and Liusie, Adian and Gales, Mark},
  booktitle={Proceedings of the 2023 conference on empirical methods in natural language processing},
  pages={9004--9017},
  year={2023}
}

@article{chern2023factool,
  title={FacTool: Factuality Detection in Generative AI--A Tool Augmented Framework for Multi-Task and Multi-Domain Scenarios},
  author={Chern, I and Chern, Steffi and Chen, Shiqi and Yuan, Weizhe and Feng, Kehua and Zhou, Chunting and He, Junxian and Neubig, Graham and Liu, Pengfei and others},
  journal={arXiv preprint arXiv:2307.13528},
  year={2023}
}

@inproceedings{lei2025factcg,
  title={FactCG: Enhancing fact checkers with graph-based multi-hop data},
  author={Lei, Deren and Li, Yaxi and Li, Siyao and Hu, Mengya and Xu, Rui and Archer, Ken and Wang, Mingyu and Ching, Emily and Deng, Alex},
  booktitle={Proceedings of the 2025 Conference of the Nations of the Americas Chapter of the Association for Computational Linguistics: Human Language Technologies (Volume 1: Long Papers)},
  pages={5002--5020},
  year={2025}
}

@inproceedings{warner2024modernbert,
  title={Smarter, better, faster, longer: A modern bidirectional encoder for fast, memory efficient, and long context finetuning and inference},
  author={Warner, Benjamin and Chaffin, Antoine and Clavi{\'e}, Benjamin and Weller, Orion and Hallstr{\"o}m, Oskar and Taghadouini, Said and Gallagher, Alexis and Biswas, Raja and Ladhak, Faisal and Aarsen, Tom and others},
  booktitle={Proceedings of the 63rd Annual Meeting of the Association for Computational Linguistics (Volume 1: Long Papers)},
  pages={2526--2547},
  year={2025}
}

@inproceedings{tasksource2025modernbertnli,
    title = "tasksource: A Large Collection of {NLP} tasks with a Structured Dataset Preprocessing Framework",
    author = "Sileo, Damien",
    booktitle = "Proceedings of the 2024 Joint International Conference on Computational Linguistics, Language Resources and Evaluation (LREC-COLING 2024)",
    month = may,
    year = "2024",
    address = "Torino, Italia",
    publisher = "ELRA and ICCL",
    url = "https://aclanthology.org/2024.lrec-main.1361",
    pages = "15655--15684",
}

@article{verifastscore2025,
  title={VERIFASTSCORE: Speeding up long-form factuality evaluation},
  author={Rajendhran, Rishanth and Zadeh, Amir and Sarte, Matthew and Li, Chuan and Iyyer, Mohit},
  journal={arXiv preprint arXiv:2505.16973},
  year={2025}
}

@article{ibm2025granite,
  title={Granite guardian},
  author={Padhi, Inkit and Nagireddy, Manish and Cornacchia, Giandomenico and Chaudhury, Subhajit and Pedapati, Tejaswini and Dognin, Pierre and Murugesan, Keerthiram and Miehling, Erik and Cooper, Mart{\'\i}n Santill{\'a}n and Fraser, Kieran and others},
  journal={arXiv preprint arXiv:2412.07724},
  year={2024}
}

@inproceedings{thawani2021numeric,
  title={Representing numbers in NLP: a survey and a vision},
  author={Thawani, Avijit and Pujara, Jay and Ilievski, Filip and Szekely, Pedro},
  booktitle={Proceedings of the 2021 Conference of the North American Chapter of the Association for Computational Linguistics: Human Language Technologies},
  pages={644--656},
  year={2021}
}

@inproceedings{mishra2022numglue,
  title={NumGLUE: A suite of fundamental yet challenging mathematical reasoning tasks},
  author={Mishra, Swaroop and Mitra, Arindam and Varshney, Neeraj and Sachdeva, Bhavdeep and Clark, Peter and Baral, Chitta and Kalyan, Ashwin},
  booktitle={Proceedings of the 60th Annual Meeting of the Association for Computational Linguistics (Volume 1: Long Papers)},
  pages={3505--3523},
  year={2022}
}

@inproceedings{wadden2020scifact,
  title={Fact or fiction: Verifying scientific claims},
  author={Wadden, David and Lin, Shanchuan and Lo, Kyle and Wang, Lucy Lu and van Zuylen, Madeleine and Cohan, Arman and Hajishirzi, Hannaneh},
  booktitle={Proceedings of the 2020 Conference on Empirical Methods in Natural Language Processing (EMNLP)},
  pages={7534--7550},
  year={2020}
}

@inproceedings{wright2022scifact,
  title={Generating scientific claims for zero-shot scientific fact checking},
  author={Wright, Dustin and Wadden, David and Lo, Kyle and Kuehl, Bailey and Cohan, Arman and Augenstein, Isabelle and Wang, Lucy Lu},
  booktitle={Proceedings of the 60th Annual Meeting of the Association for Computational Linguistics (Volume 1: Long Papers)},
  pages={2448--2460},
  year={2022}
}

@article{mishra2024llamat,
  title={Foundational large language models for materials research},
  author={Mishra, Vaibhav and Singh, Somaditya and Ahlawat, Dhruv and Zaki, Mohd and Bihani, Vaibhav and Grover, Hargun Singh and Mishra, Biswajit and Miret, Santiago and Krishnan, NM and others},
  journal={arXiv preprint arXiv:2412.09560},
  year={2024}
}

@misc{quantulum3,
  title        = {quantulum3: A {P}ython Library for Quantity Extraction from Unstructured Text},
  author       = {{The quantulum3 contributors}},
  year         = {2024},
  howpublished = {\url{https://github.com/nielstron/quantulum3}}
}

@misc{pint,
  title={{Pint}: A {P}ython Units Library},
  author={Grecco, Hern{\'a}n E and others},
  year={2024},
  howpublished={\url{https://github.com/hgrecco/pint}}
}

@article{krippendorff1980content,
  title={Content analysis: An introduction to its methodology},
  author={Ford, John M},
  journal={Personnel psychology},
  volume={57},
  number={4},
  pages={1110},
  year={2004},
  publisher={Blackwell Publishing Ltd.}
}

@inproceedings{wan2024acueval,
  title={Acueval: Fine-grained hallucination evaluation and correction for abstractive summarization},
  author={Wan, David and Sinha, Koustuv and Iyer, Srini and Celikyilmaz, Asli and Bansal, Mohit and Pasunuru, Ramakanth},
  booktitle={Findings of the Association for Computational Linguistics: ACL 2024},
  pages={10036--10056},
  year={2024}
}

@inproceedings{song2024finesure,
  title={FineSurE: Fine-grained summarization evaluation using LLMs},
  author={Song, Hwanjun and Su, Hang and Shalyminov, Igor and Cai, Jason and Mansour, Saab},
  booktitle={Proceedings of the 62nd Annual Meeting of the Association for Computational Linguistics (Volume 1: Long Papers)},
  pages={906--922},
  year={2024}
}

@inproceedings{bang2025hallulens,
  title={Hallulens: Llm hallucination benchmark},
  author={Bang, Yejin and Ji, Ziwei and Schelten, Alan and Hartshorn, Anthony and Fowler, Tara and Zhang, Cheng and Cancedda, Nicola and Fung, Pascale},
  booktitle={Proceedings of the 63rd Annual Meeting of the Association for Computational Linguistics (Volume 1: Long Papers)},
  pages={24128--24156},
  year={2025}
}

@inproceedings{vazquez2025mushroom,
  title={SemEval-2025 task 3: Mu-SHROOM, the multilingual shared-task on hallucinations and related observable overgeneration mistakes},
  author={V{\'a}zquez, Ra{\'u}l and Mickus, Timothee and Zosa, Elaine and Vahtola, Teemu and Tiedemann, J{\"o}rg and Sinha, Aman and Segonne, Vincent and S{\'a}nchez-Vega, Fernando and Raganato, Alessandro and Libovick{\`y}, Jind{\v{r}}ich and others},
  booktitle={Proceedings of the 19th International Workshop on Semantic Evaluation (SemEval-2025)},
  pages={2472--2497},
  year={2025}
}

@article{venktesh2024quantemp,
  title={Quantemp: A real-world open-domain benchmark for fact-checking numerical claims},
  author={Anand, Abhijit and Anand, Avishek and Setty, Vinay and others},
  journal={arXiv preprint arXiv:2403.17169},
  year={2024}
}

@article{venktesh2025quantempplus,
  title={A Benchmark for Open-Domain Numerical Fact-Checking Enhanced by Claim Decomposition},
  author={Venktesh, V and Prabhu, Deepali and Anand, Avishek},
  journal={arXiv preprint arXiv:2510.22055},
  year={2025}
}

@article{venktesh2025checkthat,
  title={Overview of the CLEF-2025 CheckThat! lab task 3 on fact-checking numerical claims},
  author={Venktesh, V and Setty, V and Anand, A and Hasanain, M and Bendou, B and Bouamor, H and Alam, F and Iturra-Bocaz, G and Galu{\v{s}}{\v{c}}{\'a}kov{\'a}, P},
  journal={Working Notes of CLEF},
  year={2025}
}

@article{heil2025dsgt,
  title={DS@ GT at CheckThat! 2025: evaluating context and tokenization strategies for numerical fact verification},
  author={Heil, Maximilian and Pramov, Aleksandar},
  journal={arXiv preprint arXiv:2507.06195},
  year={2025}
}

@article{anik2025claimiq,
  title={ClaimIQ at CheckThat! 2025: comparing prompted and fine-tuned language models for verifying numerical claims},
  author={Anik, Anirban Saha and Chowdhury, Md Fahimul Kabir and Wyckoff, Andrew and Choudhury, Sagnik Ray},
  journal={arXiv preprint arXiv:2509.11492},
  year={2025}
}

@inproceedings{singh2025verifierfc,
  title={Think Right, Not More: Test-Time Scaling for Numerical Claim Verification},
  author={Chungkham, Primakov and Viswanathan, Venktesh and Setty, Vinay and Anand, Avishek},
  booktitle={Empirical Methods in Natural Language Processing (EMNLP) 2025},
  pages={24345--24363},
  year={2025},
  organization={Association for Computational Linguistics}
}

@inproceedings{wadden2021sciver,
  title={Overview and insights from the SCIVER shared task on scientific claim verification},
  author={Wadden, David and Lo, Kyle},
  booktitle={Proceedings of the Second Workshop on Scholarly Document Processing},
  pages={124--129},
  year={2021}
}

@article{hallumat2025,
  title={HalluMat: Detecting Hallucinations in LLM-Generated Materials Science Content Through Multi-Stage Verification},
  author={Vangala, Bhanu Prakash and Mahmud, Sajid and Neupane, Pawan and Selvaraj, Joel and Cheng, Jianlin},
  journal={arXiv preprint arXiv:2512.22396},
  year={2025}
}

@article{medragchecker2026,
  title={MedRAGChecker: Claim-Level Verification for Biomedical Retrieval-Augmented Generation},
  author={Ji, Yuelyu and Kwak, Min Gu and Zhang, Hang and Wu, Xizhi and Li, Chenyu and Wang, Yanshan},
  journal={arXiv preprint arXiv:2601.06519},
  year={2026}
}

@article{aly2024tabver,
  title={Tabver: Tabular fact verification with natural logic},
  author={Aly, Rami and Vlachos, Andreas},
  journal={Transactions of the Association for Computational Linguistics},
  volume={12},
  pages={1648--1671},
  year={2024},
  publisher={MIT Press 255 Main Street, 9th Floor, Cambridge, Massachusetts 02142, USA~…}
}

@article{cosineverifier2025,
  title={CoSineVerifier: Tool-Augmented Answer Verification for Computation-Oriented Scientific Questions},
  author={Feng, Ruixiang and An, Zhenwei and Wen, Yuntao and Le, Ran and Jia, Yiming and Yang, Chen and Chen, Zongchao and Chen, Lisi and Gao, Shen and Shang, Shuo and others},
  journal={arXiv preprint arXiv:2512.01224},
  year={2025}
}

@inproceedings{zhu2023explain,
  title={Explain, edit, generate: rationale-sensitive counterfactual data augmentation for multi-hop fact verification},
  author={Zhu, Yingjie and Si, Jiasheng and Zhao, Yibo and Zhu, Haiyang and Zhou, Deyu and He, Yulan},
  booktitle={Proceedings of the 2023 Conference on Empirical Methods in Natural Language Processing},
  pages={13377--13392},
  year={2023}
}

@inproceedings{whitehouse2023llm,
  title={LLM-powered data augmentation for enhanced cross-lingual performance},
  author={Whitehouse, Chenxi and Choudhury, Monojit and Aji, Alham Fikri},
  booktitle={Proceedings of the 2023 conference on empirical methods in natural language processing},
  pages={671--686},
  year={2023}
}

@article{feng2021empirical,
  title={An empirical survey of data augmentation for limited data learning in nlp},
  author={Chen, Jiaao and Tam, Derek and Raffel, Colin and Bansal, Mohit and Yang, Diyi},
  journal={Transactions of the Association for Computational Linguistics},
  volume={11},
  pages={191--211},
  year={2023},
  publisher={MIT Press One Broadway, 12th Floor, Cambridge, Massachusetts 02142, USA~…}
}

@article{belinkov2022probing,
  title={Probing classifiers: Promises, shortcomings, and advances},
  author={Belinkov, Yonatan},
  journal={Computational Linguistics},
  volume={48},
  number={1},
  pages={207--219},
  year={2022}
}

@inproceedings{ribeiro2020checklist,
  title={Beyond accuracy: Behavioral testing of NLP models with CheckList},
  author={Ribeiro, Marco Tulio and Wu, Tongshuang and Guestrin, Carlos and Singh, Sameer},
  booktitle={Proceedings of the 58th annual meeting of the association for computational linguistics},
  pages={4902--4912},
  year={2020}
}

@article{rogers2020primer,
  title={A primer in BERTology: What we know about how BERT works},
  author={Rogers, Anna and Kovaleva, Olga and Rumshisky, Anna},
  journal={Transactions of the association for computational linguistics},
  volume={8},
  pages={842--866},
  year={2020}
}

@inproceedings{zhang2025dually,
  title={Dually Self-Improved Counterfactual Data Augmentation Using Large Language Model},
  author={Zhang, Luhao and Zhang, Xinyu and Hu, Linmei and Song, Dandan and Nie, Liqiang},
  booktitle={Proceedings of the 63rd Annual Meeting of the Association for Computational Linguistics (Volume 1: Long Papers)},
  pages={5216--5227},
  year={2025}
}

@inproceedings{mi2025blindspots,
  title={From Input Perception to Predictive Insight: Modeling Model Blind Spots Before They Become Errors},
  author={Mi, Maggie and Villavicencio, Aline and Moosavi, Nafise Sadat},
  booktitle={Proceedings of the 2025 Conference on Empirical Methods in Natural Language Processing},
  pages={34316--34329},
  year={2025}
}

@inproceedings{conll2025ood,
  title={Beyond accuracy: Revisiting out-of-distribution generalization in NLI models},
  author={Delbari, Zahra and Pilehvar, Mohammad Taher},
  booktitle={Proceedings of the 29th Conference on Computational Natural Language Learning},
  pages={557--570},
  year={2025}
}

@article{seo2025verifying,
  title={Verifying the verifiers: Unveiling pitfalls and potentials in fact verifiers},
  author={Seo, Wooseok and Han, Seungju and Jung, Jaehun and Newman, Benjamin and Lim, Seungwon and Lee, Seungbeen and Lu, Ximing and Choi, Yejin and Yu, Youngjae},
  journal={arXiv preprint arXiv:2506.13342},
  year={2025}
}
